\pdfoutput=1
\documentclass{article} %
\usepackage{iclr2027_conference,times}

\usepackage{amsmath,amsfonts,bm}

\def\eqref#1{equation~\ref{#1}}
\def\1{\bm{1}}

\DeclareMathAlphabet{\mathsfit}{\encodingdefault}{\sfdefault}{m}{sl}
\SetMathAlphabet{\mathsfit}{bold}{\encodingdefault}{\sfdefault}{bx}{n}

\usepackage{hyperref}
\usepackage{url}
\usepackage{booktabs}
\usepackage{multirow}
\usepackage{graphicx}
\usepackage{amsmath,amssymb}
\usepackage{amsthm}
\usepackage{algorithm}
\usepackage{algpseudocode}

\renewcommand{\eqref}[1]{Equation~(\ref{#1})}

\newtheorem{theorem}{Theorem}[section]

\theoremstyle{remark}

\usepackage{xcolor}
\usepackage{colortbl}
\usepackage{arydshln}
\usepackage{subcaption}
\usepackage{wrapfig}

\definecolor{oursrowtint}{RGB}{234,239,249}
\newcommand{\ours}{\rowcolor{oursrowtint}}

\newcommand{\todocell}[1]{\textcolor{gray}{--}}

\title{RAQ-MLA: Role-Aware Quantization for\\
Multi-Head Latent Attention Caches}

\author{
Tianqi Chen$^{1,3,*}$,\;
Xunchao Li$^{2,*}$,\;
Linchuan Xie$^{2}$,\;
Guanghua Yu$^{2}$ \\
\bf Jianchen Zhu$^{2}$,\;
Peisong Wang$^{1,3}$,\;
Jian Cheng$^{1,3}$\,$^{\dagger}$
\\[4pt]
$^{1}$C$^{2}$DL, Institute of Automation, Chinese Academy of Sciences
\quad $^{2}$Tencent \\
$^{3}$School of Artificial Intelligence, University of Chinese Academy of Sciences \\[3pt]
\texttt{chentianqi2023@ia.ac.cn}, \texttt{lucayu@tencent.com},
\texttt{jcheng@nlpr.ia.ac.cn} \\[2pt]
$^{*}$Equal contribution.\quad $^{\dagger}$Corresponding author.
}

\iclrfinalcopy
\begin{document}

\maketitle
\lhead{Preprint}

\begin{abstract}
Multi-Head Latent Attention (MLA) has been adopted by recent frontier large
language models. Instead of per-head keys and values, MLA caches a compressed
KV latent that is up-projected into keys and values at decode time, together
with a narrow positional key that carries position information into the
attention score. Existing KV-cache quantization methods are designed around the
keys and values of standard attention and do not account for what MLA actually
caches, which costs accuracy at low bit widths. We analyze the MLA cache both
in activation distribution and in how quantization error propagates to the
attention output, and show that the two cached tensors play different roles:
the latent mainly supplies the values, whereas the positional key dominates the
attention score. We therefore propose \textbf{RAQ-MLA} (Role-Aware Quantization
for Multi-Head Latent Attention Caches), which fits an invertible transform to
each cached tensor against the error that tensor delivers to the attention
output: attention-weighted value error for the latent and score perturbation for
the positional key. Without training, RAQ-MLA quantizes the full MLA cache to
about $2.5$ bits per element, a $6.5\times$ increase in theoretical KV-cache
capacity, while remaining nearly lossless on reasoning and coding tasks across
four MLA models.
\end{abstract}

\section{Introduction}
\label{sec:intro}

Long-horizon agents have made ultra-long-context inference a mainstream serving
workload \citep{yao2023react,yang2024sweagentagentcomputerinterfacesenable,deepseekharness}.
Serving it requires more than efficient long-sequence computation: the
key-value (KV) cache must be held for the entire lifetime of a request and grows
linearly with both context length and concurrency. At runtime, the cache
competes with batch size for HBM capacity \citep{pagedattention}; when persisted
for prefix reuse, it consumes host memory, SSD capacity, and interconnect
bandwidth \citep{mooncake,sglang}. Both pressures cap serving throughput and
raise deployment cost, and both grow as agents run longer and at higher
concurrency.

To reduce this footprint, Multi-Head Latent Attention (MLA) \citep{deepseekv2}
compresses each token's key and value into a single low-dimensional latent
$\mathbf{c}^{\mathrm{KV}}_t$, which is cached and up-projected into keys and values at
decode time, alongside a narrow decoupled positional key $\mathbf{k}^{\mathrm{R}}_t$.
MLA has been adopted by recent frontier models, increasingly together with
sparse attention \citep{deepseekv32,glm5}.
Sparse attention, however, only selects which cached tokens each query attends
to; the full cache is still retained. MLA thus shrinks the cache per token, but
neither it nor sparse attention changes its linear growth with context length:
at the one-million-token context
length of recent models, a single sequence still occupies tens of GiB of cache.
Compressing what MLA actually stores therefore remains necessary.

KV-cache quantization is the other line of work that reduces cache size,
lowering the average bits per stored element without any training
\citep{kvquant,kivi,oscar,rotatekv}. However, most prior methods
\citep{kivi,kvquant} design their granularity, precision, and transforms around
the keys and values of standard attention. MLA caches neither: it stores the
compressed latent $\mathbf{c}^{\mathrm{KV}}_t$ and the decoupled positional key
$\mathbf{k}^{\mathrm{R}}_t$, and recovers the content key and value from the latent only
at use time. Existing MLA cache quantizers \citep{snapmla,kvarn} keep the
latent at four or more bits and leave $\mathbf{k}^{\mathrm{R}}$ in full precision. In
this paper, we aim to quantize the full MLA cache, including the positional key,
to extremely low bit widths (about $2.5$ bits per cached element on average).

\begin{figure}[t]
\centering
\includegraphics[width=\textwidth]{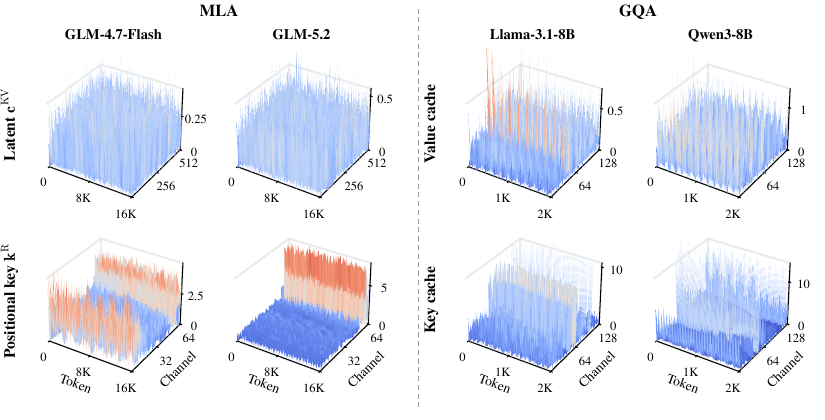}
\caption{\textbf{Fixed-channel outliers appear in the decoupled positional key
but not in the compressed KV latent.} The positional key behaves like the Key
cache of standard attention, and the compressed latent like the Value cache.}
\label{fig:outlier-dist}
\end{figure}

We analyze the two cached tensors both in distribution and in how their
quantization error propagates through the MLA computation. Fixed-channel
outliers appear in the decoupled positional key but not in the compressed
latent (Figure~\ref{fig:outlier-dist}), and the positional key carries
$92.5\%$ of the attention-score attribution, most of it in a few outlier
channels. As a result, the two tensors reach the output through different
routes. At $4$ bits, the latent's $12.3\%$ error in the cache falls to $1.7\%$
at the attention output through the content key but remains $8.3\%$ through the
value, whereas the positional key's error rises from $9.1\%$ to $14.7\%$. The
latent therefore mainly plays the value role in MLA, while the tensor that sets
the score is the one existing MLA methods exempt from quantization.

\begin{wrapfigure}[17]{r}{0.44\textwidth}
\vspace{-\intextsep}
\centering
\includegraphics[width=\linewidth]{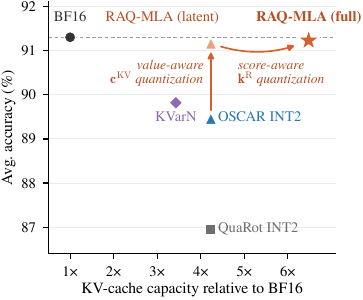}
\caption{Cache capacity against accuracy on GLM-4.7-Flash.}
\label{fig:capacity-accuracy}
\end{wrapfigure}

Motivated by this, we propose RAQ-MLA (Role-Aware Quantization for Multi-Head
Latent Attention Caches). It quantizes the latent per token and the positional
key per channel, following the distribution each exhibits, and stores each in
an invertible basis fitted against the error that tensor delivers to the
attention output. Both bases are fitted under a single objective, a quadratic
form weighted per token by a \emph{role metric} built from how attention
consumes the tensor: for the latent, the value up-projection weighted by how
strongly each token is attended; for the positional key, the directions of the
queries that read it. We solve it by alternating minimization, requiring
neither gradients through the model nor fine-tuning. Both transforms stay inside
the compact spaces MLA already works in, $512\times512$ for the latent and
$64\times64$ for the positional key, so RAQ-MLA improves low-bit accuracy at
negligible extra computation.

Our contributions are:

\begin{itemize}
\item \textbf{Role analysis of the MLA cache.} We show that the two cached
tensors play different roles in attention: error in the compressed latent
reaches the output mainly through the value readout, whereas the decoupled
positional key dominates the attention score and amplifies its own
quantization error (Section~\ref{sec:motivation}).
\item \textbf{Role-aware quantization.} We propose RAQ-MLA, a training-free
quantization method designed for the MLA cache. It pairs per-token quantization
with a value-aware transform for the latent, and per-channel quantization with
a score-aware transform for the positional key. Both transforms are fitted under
a single role-aware objective by alternating minimization
(Section~\ref{sec:method}).
\item \textbf{Full-cache quantization at about 2.5 bits.} On four MLA models
from the GLM, DeepSeek, and Moonlight families, RAQ-MLA quantizes both cached
tensors to about $2.5$ bits per element, a $6.5\times$ increase in theoretical
KV-cache capacity, while staying within $0.5$ points of BF16 on reasoning and
coding (Figure~\ref{fig:capacity-accuracy}).
\end{itemize}

\section{Background: Multi-Head Latent Attention}
\label{sec:background}

Multi-Head Latent Attention (MLA) \citep{deepseekv2} was introduced to shrink
the KV cache of multi-head attention; we briefly recall its computation and the
cache it stores. Per-token activations are row vectors, a layer has $n_{\mathrm{h}}$
heads of width $d_{\mathrm{h}}$ (used for both the content key and the value for
simplicity), and $\mathbf{h}_{t}$ is the attention input at token $t$.

\paragraph{Low-rank key-value joint compression.}
MLA compresses keys and values into a shared latent and recovers content
(non-positional) keys and values by per-head up-projections:
\begin{equation}
\mathbf{c}_{t}^{\mathrm{KV}} = \mathrm{RMSNorm}\bigl(\mathbf{h}_{t}(\mathbf{W}^{\mathrm{DKV}})^{\top}\bigr),
\qquad
\mathbf{k}_{t,h}^{\mathrm{C}} = \mathbf{c}_{t}^{\mathrm{KV}}(\mathbf{W}_{h}^{\mathrm{UK}})^{\top},
\qquad
\mathbf{v}_{t,h}^{\mathrm{C}} = \mathbf{c}_{t}^{\mathrm{KV}}(\mathbf{W}_{h}^{\mathrm{UV}})^{\top},
\label{eq:mla-compress}
\end{equation}
where $\mathbf{c}_{t}^{\mathrm{KV}}\!\in\!\mathbb{R}^{d_{\mathrm{c}}}$ is the compressed KV latent
with $d_{\mathrm{c}}\!\ll\!n_{\mathrm{h}}d_{\mathrm{h}}$, and
$\mathbf{W}_{h}^{\mathrm{UK}},\mathbf{W}_{h}^{\mathrm{UV}}\!\in\!\mathbb{R}^{d_{\mathrm{h}}\times d_{\mathrm{c}}}$ are the head-$h$
blocks of the up-projections. $\mathbf{k}_{t,h}^{\mathrm{C}}$ and $\mathbf{v}_{t,h}^{\mathrm{C}}$
are not cached: at inference, $\mathbf{W}_{h}^{\mathrm{UK}}$ is absorbed into the query side and
$\mathbf{W}_{h}^{\mathrm{UV}}$ into the output side, so neither is ever materialized.

\paragraph{Decoupled rotary position embedding.}
MLA utilizes a \emph{decoupled RoPE} strategy to carry position information:
per-head positional queries $\mathbf{q}_{t,h}^{\mathrm{R}}\!\in\!\mathbb{R}^{d^{\mathrm{R}}}$
and a single positional key $\mathbf{k}_{t}^{\mathrm{R}}\!\in\!\mathbb{R}^{d^{\mathrm{R}}}$
\emph{shared by all $n_{\mathrm{h}}$ heads} are rotated by RoPE, while
$\mathbf{k}_{t,h}^{\mathrm{C}}$ remains the content key. With a low-rank query latent
$\mathbf{c}_{t}^{\mathrm{Q}}$,
\begin{align}
\mathbf{c}_{t}^{\mathrm{Q}} &= \mathrm{RMSNorm}\bigl(\mathbf{h}_{t}(\mathbf{W}^{\mathrm{DQ}})^{\top}\bigr),
&
\mathbf{q}_{t,h}^{\mathrm{C}} &= \mathbf{c}_{t}^{\mathrm{Q}}(\mathbf{W}_{h}^{\mathrm{UQ}})^{\top},
\label{eq:mla-query}\\
\mathbf{q}_{t,h}^{\mathrm{R}} &= \mathrm{RoPE}\bigl(\mathbf{c}_{t}^{\mathrm{Q}}(\mathbf{W}_{h}^{\mathrm{QR}})^{\top}\bigr),
&
\mathbf{k}_{t}^{\mathrm{R}} &= \mathrm{RoPE}\bigl(\mathbf{h}_{t}(\mathbf{W}^{\mathrm{KR}})^{\top}\bigr),
\label{eq:mla-rope}\\
\mathbf{q}_{t,h} &= \bigl[\mathbf{q}_{t,h}^{\mathrm{C}};\,\mathbf{q}_{t,h}^{\mathrm{R}}\bigr],
&
\mathbf{k}_{t,h} &= \bigl[\mathbf{k}_{t,h}^{\mathrm{C}};\,\mathbf{k}_{t}^{\mathrm{R}}\bigr],
\label{eq:mla-concat}
\end{align}
where $[\cdot;\cdot]$ denotes concatenation.

\paragraph{Attention computation.}
With the concatenated queries and keys, MLA computes
\begin{equation}
\mathbf{o}_{q,h}=\sum_{t=1}^{q} S_{h,q,t}\,\mathbf{v}_{t,h}^{\mathrm{C}},
\qquad
S_{h,q,t}=\mathrm{Softmax}_{t}\!\left(
\frac{\mathbf{q}_{q,h}^{\mathrm{C}}(\mathbf{k}_{t,h}^{\mathrm{C}})^{\top}
+\mathbf{q}_{q,h}^{\mathrm{R}}(\mathbf{k}_{t}^{\mathrm{R}})^{\top}}{\sqrt{d_{\mathrm{h}}+d^{\mathrm{R}}}}
\right),
\label{eq:mla-attention}
\end{equation}
where $S_{h,q,t}$ is the weight head $h$ assigns to cached token $t$ when
answering the query at position $q$, and the logit splits into a content term
and a positional term. The layer output is assembled by the usual output
projection over heads.

\paragraph{Inference cache.}
The persisted cache therefore holds two tensors per layer: the compressed
latent $\mathbf{c}^{\mathrm{KV}}_{t}$ of width $d_{\mathrm{c}}$, and the decoupled positional key
$\mathbf{k}^{\mathrm{R}}_{t}$ of width $d^{\mathrm{R}}$, shared across heads. Most MLA models set
$d_{\mathrm{c}}=512$ and $d^{\mathrm{R}}=64$
\citep{deepseekv2,deepseekcoderv2,deepseekv32,moonlight,glm5}, so the positional
key accounts for $64/576=11.1\%$ of the cached elements.

\section{Motivation}
\label{sec:motivation}

We analyze the MLA cache from a quantization perspective to decide, for each
cached tensor, how it should be quantized and what its quantized bits must
preserve. The analysis proceeds in three steps: the distribution of each
tensor, how its quantization error propagates to the attention output, and its
contribution to the attention score, which explains that propagation. Unless
stated otherwise, all analyses use GLM-4.7-Flash on calibration prompts.

\begin{figure}[t]
\centering
\begin{subfigure}[b]{0.495\textwidth}
  \centering
  \includegraphics[width=\textwidth]{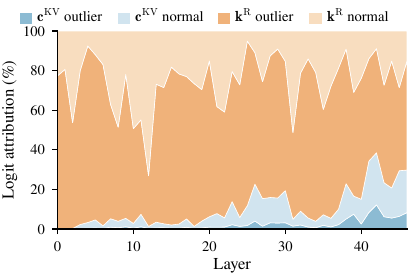}
  \caption{}
  \label{fig:score-attribution}
\end{subfigure}
\hfill
\begin{subfigure}[b]{0.495\textwidth}
  \centering
  \includegraphics[width=\textwidth]{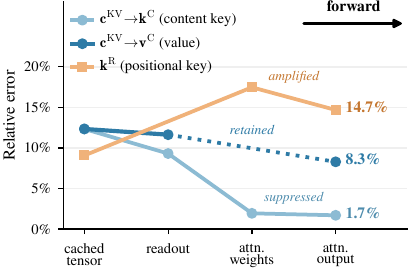}
  \caption{}
  \label{fig:error-fate}
\end{subfigure}
\caption{\textbf{The positional key dominates the attention score, while the
latent mainly plays the value role.}
(a) Logit attribution to the outlier and normal channels of the latent
$\mathbf{c}^{\mathrm{KV}}$ and the positional key $\mathbf{k}^{\mathrm{R}}$, per layer.
(b) Relative error at each stage under INT4, with the latent quantized per token
and the positional key per channel.}
\label{fig:kr-evidence}
\end{figure}

\paragraph{How are the two cached tensors distributed?}
As shown in Figure~\ref{fig:outlier-dist}, channel structure separates the two
cached tensors. We compare $\mathbf{c}^{\mathrm{KV}}$ and $\mathbf{k}^{\mathrm{R}}$ in
GLM-4.7-Flash and GLM-5.2 with the Key and Value caches of standard (non-MLA)
attention in Llama-3.1-8B \citep{llama3} and Qwen3-8B \citep{qwen3}, and each
aligns with one of the two references. $\mathbf{k}^{\mathrm{R}}$ carries the
channel-aligned outliers that conventional Key caches exhibit.
$\mathbf{c}^{\mathrm{KV}}$ instead resembles a Value cache, spreading its magnitude
broadly, with no fixed channel standing out. The first design decision in
RAQ-MLA follows: the positional key is quantized per channel, which keeps its
outlier channels out of a shared scale, and the latent per token, which
confines the error to the token being quantized \citep{kivi}. This granularity
reduces the quantization error of each cached tensor, but it does not target
the errors that matter at the attention output.

\paragraph{How does quantization error reach the output?}
We quantize the latent per token and the positional key per channel, both at
4 bits, and measure the relative error stage by stage
(Figure~\ref{fig:error-fate}). The latent's two readouts start out alike: right
after up-projection, $\mathbf{k}^{\mathrm{C}}$ and $\mathbf{v}^{\mathrm{C}}$ carry comparable
relative error. Only the content key enters the attention weights, and there
its error has shrunk to $1.9\%$, an order of magnitude below the positional
key's $17.5\%$. At the attention output, the latent's $12.3\%$ error in the cache has
fallen to $1.7\%$ through the content key but remains $8.3\%$ through the
value, whereas the positional key's error has risen from $9.1\%$ to $14.7\%$.

\paragraph{Why do the two routes differ?}
The channel structure of Figure~\ref{fig:outlier-dist} accounts for this split.
The outlier channels of $\mathbf{k}^{\mathrm{R}}$ dominate the attention score: pooled
over all $47$ layers, the positional term carries $92.5\%$ of the logit
attribution (Appendix~\ref{app:logit-attribution}) against $7.5\%$ for the
content term, and the outlier channels of $\mathbf{k}^{\mathrm{R}}$, a tenth of
its $64$, carry $63.8\%$ on their own (Figure~\ref{fig:score-attribution}).
Because the content term contributes so
little to the logit, perturbing it barely moves the attention weights; the
positional term sets them, so its error is amplified into the attention weights
and carried on to the output. Within the latent, then, the value readout
carries most of its error through to the output, while the attention score is
dominated by $\mathbf{k}^{\mathrm{R}}$.

\paragraph{Why do we choose a learnable transform for the MLA cache?}
These roles call for a storage basis chosen by how attention reads each tensor
rather than by its distribution: the latent's error should stay out of the
directions the value up-projection reads, and the positional key's out of the
directions its queries probe. We therefore propose RAQ-MLA, which learns a
basis for each cached tensor against the error that tensor delivers to the
attention output. The MLA architecture keeps such transforms cheap. First, both cached tensors are low-dimensional ($d_{\mathrm{c}}=512$ and $d^{\mathrm{R}}=64$), so
the two transforms together cost $d_{\mathrm{c}}^2+(d^{\mathrm{R}})^2$ coefficients per layer
($0.08\%$ of the total parameters of GLM-4.7-Flash). Second, as detailed in
Section~\ref{sec:method-serving}, the cached tensors never have to be
transformed back at decode time: for the latent, the inverse folds into
$\mathbf{W}^{\mathrm{UK}}$ and $\mathbf{W}^{\mathrm{UV}}$; for the positional key, the compensation is applied to
the query $\mathbf{q}^{\mathrm{R}}$, which is never cached, and cancels the transform in
the attention score.

\section{RAQ-MLA}
\label{sec:method}

\begin{figure}[t]
\centering
\includegraphics[width=\textwidth]{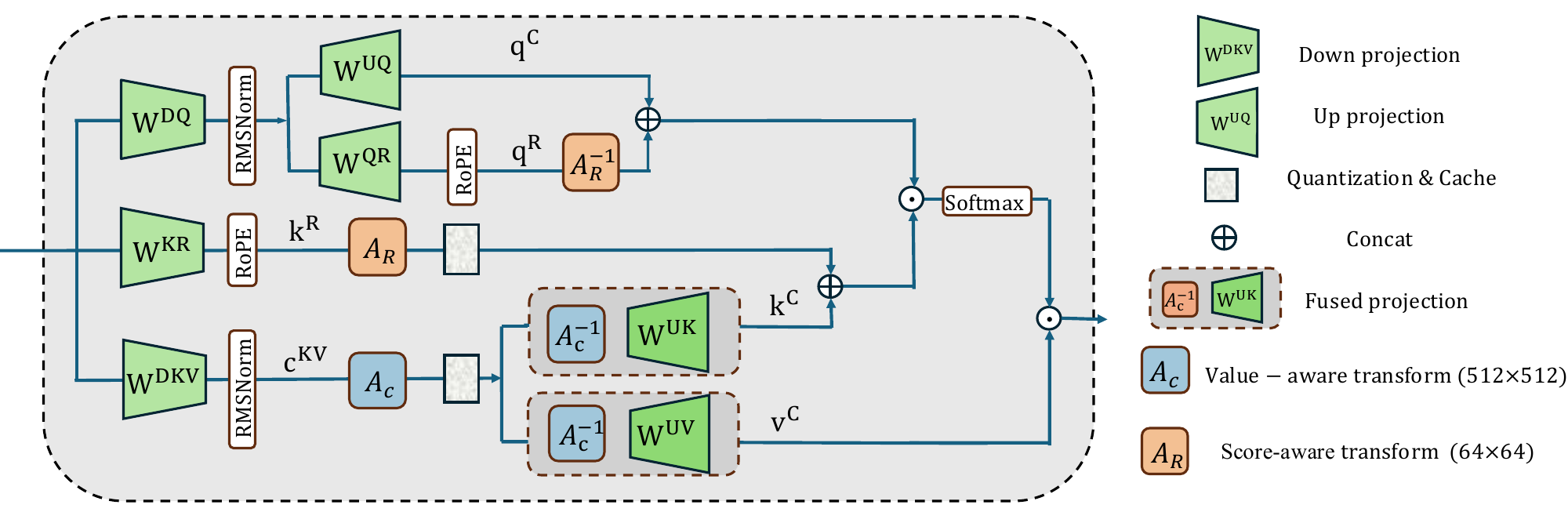}
\caption{\textbf{RAQ-MLA at one MLA layer.} $A_{\mathrm{c}}$ is applied to the latent
after RMSNorm and $A_{\mathrm{R}}$ to the positional key after RoPE, so
both cached tensors are quantized in a learned basis. On the read path
$A_{\mathrm{c}}^{-1}$ is fused into $\mathbf{W}^{\mathrm{UK}}$ and $\mathbf{W}^{\mathrm{UV}}$, while the positional key is
compensated on the query side.}
\label{fig:method-pipeline}
\end{figure}

\subsection{Formulation}
\label{sec:method-formulation}

RAQ-MLA quantizes the latent and the positional key in learned invertible bases
$A_{\mathrm{c}}$ and $A_{\mathrm{R}}$ (Figure~\ref{fig:method-pipeline}), both fitted under one
objective. We state it once for a generic cached tensor; a separate transform is
fitted for each cached tensor at every layer. Write
$X\in\mathbb{R}^{T\times d}$ for a cached tensor with rows $\mathbf{x}_t$, let
$\mathcal{Q}$ be the quantize--dequantize operator, and let $A$ denote its
transform ($A_{\mathrm{c}}$ or $A_{\mathrm{R}}$), so that the dequantized tensor is
$\widehat X=\mathcal{Q}(XA)A^{-1}$. Without quantization $\widehat X=X$, so $A$
changes only how the quantization error is distributed, not the model. We fit
$A$ by minimizing
\begin{equation}
F(A):=\mathcal{D}\!\left(\mathcal{Q}(XA)A^{-1}\right),
\qquad
\mathcal{D}(\widehat X)
=\sum_t(\widehat{\mathbf{x}}_t-\mathbf{x}_t)M_t
(\widehat{\mathbf{x}}_t-\mathbf{x}_t)^\top,
\qquad M_t\succeq0,
\label{eq:general-objective}
\end{equation}
a token-weighted quadratic in which the \emph{role metric}
$M_t\in\mathbb{R}^{d\times d}$ encodes how attention reads the tensor at token
$t$: the same error costs more in directions attention reads strongly.
$M_t$ plays the role of the Hessian in second-order post-training quantization
objectives \citep{obc,gptq}, with one difference: a weight matrix has a single
Hessian per layer, whereas $M_t$ comes from how attention consumes the cached
tensor and therefore varies from token to token.
Sections~\ref{sec:method-latent} and~\ref{sec:method-kR} derive $M_t$ for the
two cached tensors from calibration statistics, together with the
approximations each involves.

\subsection{Value-aware quantization of the compressed KV latent}
\label{sec:method-latent}

Let $C^{\mathrm{KV}}\in\mathbb{R}^{T\times d_{\mathrm{c}}}$ collect the cached latent vectors. The
latent is quantized as $Z_{\mathrm{c}}=\mathcal{Q}(C^{\mathrm{KV}}A_{\mathrm{c}})$ and read as
$\widehat{C}^{\mathrm{KV}}=Z_{\mathrm{c}}A_{\mathrm{c}}^{-1}$, with $A_{\mathrm{c}}^{-1}$ folded into the up-projections
at inference (Section~\ref{sec:method-serving}).

\paragraph{Value-aware objective.}
We model the latent's distortion along the value path and neglect the
content-key path: as demonstrated in Section~\ref{sec:motivation}, the latent's
error reaches the attention output as $1.7\%$ through the content key against
$8.3\%$ through the value. Write
$\mathbf{e}_t=\widehat{\mathbf{c}}^{\mathrm{KV}}_t-\mathbf{c}^{\mathrm{KV}}_t$ for the latent's
quantization error at token $t$; it perturbs the value that head $h$ reads from
token $t$ by $\Delta\mathbf{v}^{\mathrm{C}}_{t,h}=\mathbf{e}_t(\mathbf{W}_{h}^{\mathrm{UV}})^{\top}$. Let
$S_h\in\mathbb{R}^{T\times T}$ collect the causal attention weights
$S_{h,q,t}$ of head $h$ from \eqref{eq:mla-attention}, held at the values the
unquantized model produces; holding $S_h$ fixed is exactly what sets the
content-key path aside. The attention output of query $q$ at head $h$ is then
perturbed by $\Delta\mathbf{o}_{q,h}=\sum_t S_{h,q,t}\Delta\mathbf{v}^{\mathrm{C}}_{t,h}$,
and summing over queries and heads gives
\begin{equation}
\sum_{h,q}\left\|\Delta\mathbf{o}_{q,h}\right\|_2^2
=
\sum_{h,t,t'}[S_h^\top S_h]_{t,t'}
\langle\Delta\mathbf{v}^{\mathrm{C}}_{t,h},\Delta\mathbf{v}^{\mathrm{C}}_{t',h}\rangle .
\end{equation}
Because the latent is quantized per token, errors at different tokens are
nearly uncorrelated (Appendix Figure~\ref{fig:app-value-crosstoken}), so each
off-diagonal term is small. We therefore keep only the diagonal
$\gamma_{t,h}=[S_h^\top S_h]_{t,t}=\sum_q S_{h,q,t}^2$, which we call the
\emph{attention exposure} of token $t$ at head $h$: how strongly head $h$ reads
token $t$ over all queries. This puts the value-path distortion in the form of
\eqref{eq:general-objective},
\begin{align}
\mathcal{D}_{\mathrm{c}}(\widehat{C}^{\mathrm{KV}})
&:=\sum_{h,t}\gamma_{t,h}
\left\|\Delta\mathbf{v}^{\mathrm{C}}_{t,h}\right\|_2^2
=\sum_t\mathbf{e}_tM^{\mathrm{c}}_t\mathbf{e}_t^\top ,
\label{eq:value-distortion}\\
M^{\mathrm{c}}_t&=\sum_h\gamma_{t,h}\,(\mathbf{W}_{h}^{\mathrm{UV}})^{\top}\mathbf{W}_{h}^{\mathrm{UV}}
\ \in\mathbb{R}^{d_{\mathrm{c}}\times d_{\mathrm{c}}}.
\label{eq:latent-metric}
\end{align}

\begin{figure}[t]
\centering
\begin{subfigure}[t]{0.495\textwidth}
  \centering
  \includegraphics[width=\textwidth]{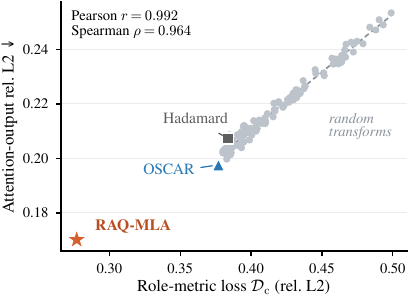}
  \caption{}
  \label{fig:role-metric-correlation}
\end{subfigure}
\hfill
\begin{subfigure}[t]{0.495\textwidth}
  \centering
  \includegraphics[width=\textwidth]{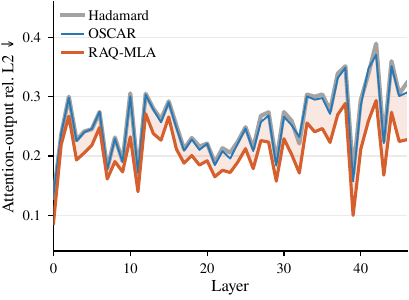}
  \caption{}
  \label{fig:role-metric-layerwise}
\end{subfigure}
\caption{\textbf{Evaluation of the latent role metric.}
(a) At layer~$23$, the role-metric loss and the exact attention-output error it
is meant to predict agree across $128$ random transforms (Pearson $r=0.992$,
Spearman $\rho=0.964$), and RAQ-MLA minimizes both.
(b) Layer-wise attention-output error of each method on held-out validation
prompts; shading marks the reduction from the better baseline to RAQ-MLA.}
\label{fig:role-metric-validation}
\end{figure}

The value up-projection sets which directions of the latent are read at all, and
$\gamma_{t,h}$ scales that geometry by how strongly head $h$ reads token $t$.
To test whether this metric tracks the error that actually reaches the output,
we score $128$ random transforms at layer~$23$ by both $\mathcal{D}_{\mathrm{c}}$ and the
exact attention-output error (Figure~\ref{fig:role-metric-correlation}). The two
agree almost linearly (Pearson $r=0.992$, Spearman $\rho=0.964$), so minimizing
$\mathcal{D}_{\mathrm{c}}$ is a faithful proxy for minimizing the attention-output error.
Fitting $A_{\mathrm{c}}$ this way lowers the attention-output error below Hadamard and
OSCAR at every one of the $47$ layers, and the reduction carries over to
held-out validation prompts (Figure~\ref{fig:role-metric-layerwise}).

\subsection{Score-aware quantization of the decoupled positional key}
\label{sec:method-kR}

Section~\ref{sec:motivation} showed that the positional key carries $92.5\%$ of
the attention-score attribution and that its error is amplified on the way to
the output, from $9.1\%$ in the cache to $14.7\%$ at the attention output. Let
$K^{\mathrm{R}}\in\mathbb{R}^{T\times d^{\mathrm{R}}}$ collect the positional keys after RoPE. The
positional key is quantized as $Z_{\mathrm{R}}=\mathcal{Q}(K^{\mathrm{R}}A_{\mathrm{R}})$ and read as
$\widehat{K}^{\mathrm{R}}=Z_{\mathrm{R}}A_{\mathrm{R}}^{-1}$; since the positional query is never cached,
$A_{\mathrm{R}}^{-1}$ is compensated on the query side at inference
(Section~\ref{sec:method-serving}).

\paragraph{Score-aware objective.}
We take the perturbation of the pre-softmax positional logit as the distortion.
Writing $\mathbf{e}^{\mathrm{R}}_t=\widehat{\mathbf{k}}^{\mathrm{R}}_t-\mathbf{k}^{\mathrm{R}}_t$, this again
gives the form of \eqref{eq:general-objective},
\begin{align}
\mathcal{D}_{\mathrm{R}}(\widehat{K}^{\mathrm{R}})
&:=\sum_{h,q}\sum_{t\le q}
\left(\mathbf{q}^{\mathrm{R}}_{q,h}(\mathbf{e}^{\mathrm{R}}_t)^{\top}\right)^2
=\sum_t\mathbf{e}^{\mathrm{R}}_tM^{\mathrm{R}}_t(\mathbf{e}^{\mathrm{R}}_t)^{\top} ,
\label{eq:score-distortion}\\
M^{\mathrm{R}}_t&=\sum_h\sum_{q\ge t}
(\mathbf{q}^{\mathrm{R}}_{q,h})^{\top}\mathbf{q}^{\mathrm{R}}_{q,h}
\ \in\mathbb{R}^{d^{\mathrm{R}}\times d^{\mathrm{R}}}.
\label{eq:key-metric}
\end{align}
$M^{\mathrm{R}}_t$ is the second moment of the positional queries that read token $t$, so
it penalizes error along the directions those queries probe most. Across $128$
random transforms at layer~$23$, $\mathcal{D}_{\mathrm{R}}$ also tracks the exact
attention-output error (Pearson $r=0.852$, Spearman $\rho=0.847$;
Appendix Figure~\ref{fig:app-kpe-role-metric}).

\subsection{Optimization}
\label{sec:method-fit}

Both objectives are instances of \eqref{eq:general-objective} and are fitted by
the same procedure. Because $\mathcal{Q}$ rounds to a grid, $F$ is
discontinuous in $A$ and the gradient through the rounding vanishes almost
everywhere, so rather than relying on straight-through estimates we alternate
between the quantized tensor and the transform. Holding
$Z^{(r)}=\mathcal{Q}(XA^{(r)})$ fixed, the distortion becomes a convex quadratic
in the inverse transform $B=A^{-1}$, which we update with a proximal term and
then invert,
\begin{equation}
A^{(r+1)}
=
\left(
\arg\min_{B}\
\mathcal{D}\!\left(Z^{(r)}B\right)
+\lambda\left\|B-(A^{(r)})^{-1}\right\|_F^2
\right)^{-1},
\label{eq:transform-update}
\end{equation}
after which $X$ is requantized in the new basis. The proximal term keeps the
update close to the current inverse transform and makes the inner problem
strongly convex for $\lambda>0$. $M_t$ is instantiated by
\eqref{eq:latent-metric} for the latent and by \eqref{eq:key-metric} for the
positional key, and no gradient is taken through the model.
Algorithm~\ref{alg:raq-mla-fit} in Appendix~\ref{app:algorithm} gives the full
procedure.

\begin{theorem}
\label{thm:descent}
Let $\lambda>0$, write $B^{(r)}=(A^{(r)})^{-1}$ and
$\Delta_r=\|B^{(r+1)}-B^{(r)}\|_F^2$, so that
$F(A^{(r)})=\mathcal{D}(Z^{(r)}B^{(r)})$ is the objective value at round $r$.
If each $B^{(r+1)}$ is invertible and each requantization does not increase the
distortion at the current decoder,
$\mathcal{D}(Z^{(r+1)}B^{(r+1)})\le\mathcal{D}(Z^{(r)}B^{(r+1)})$, then
$F(A^{(r+1)})\le F(A^{(r)})-\lambda\Delta_r$; hence $F(A^{(r)})$ converges,
$\Delta_r\to0$, and $F(A^{(R)})\le F(A^{(0)})$ for every $R$.
\end{theorem}

\subsection{Inference}
\label{sec:method-serving}

At inference, each cached tensor is written directly in its learned basis, and
its inverse transform is moved to the read side. On the write path,
\begin{equation}
\widetilde{\mathbf{c}}^{\mathrm{KV}}_t
=\mathrm{RMSNorm}\!\left(\mathbf{h}_t(\mathbf{W}^{\mathrm{DKV}})^{\top}\right)A_{\mathrm{c}},
\qquad
\widetilde{\mathbf{k}}^{\mathrm{R}}_t
=\mathrm{RoPE}\!\left(\mathbf{h}_t(\mathbf{W}^{\mathrm{KR}})^{\top}\right)A_{\mathrm{R}},
\label{eq:write-path}
\end{equation}
and the cache stores $\mathcal{Q}(\widetilde{\mathbf{c}}^{\mathrm{KV}}_t)$ and
$\mathcal{Q}(\widetilde{\mathbf{k}}^{\mathrm{R}}_t)$, i.e., the rows of $Z_{\mathrm{c}}$ and $Z_{\mathrm{R}}$.
On the read path, $A_{\mathrm{c}}^{-1}$ is folded into the up-projections and
$A_{\mathrm{R}}^{-\top}$ is applied to the positional query after RoPE,
\begin{equation}
(\widetilde{\mathbf{W}}_{h}^{\mathrm{UK}})^{\top}=A_{\mathrm{c}}^{-1}(\mathbf{W}_{h}^{\mathrm{UK}})^{\top},
\qquad
(\widetilde{\mathbf{W}}_{h}^{\mathrm{UV}})^{\top}=A_{\mathrm{c}}^{-1}(\mathbf{W}_{h}^{\mathrm{UV}})^{\top},
\qquad
\widetilde{\mathbf{q}}^{\mathrm{R}}_{q,h}=\mathbf{q}^{\mathrm{R}}_{q,h}A_{\mathrm{R}}^{-\top},
\label{eq:latent-folding}
\end{equation}
so that $Z_{\mathrm{c}}(\widetilde{\mathbf{W}}_{h}^{\mathrm{UK}})^{\top}=\widehat{C}^{\mathrm{KV}}(\mathbf{W}_{h}^{\mathrm{UK}})^{\top}$, likewise
for $\mathbf{W}_{h}^{\mathrm{UV}}$, and
$\widetilde{\mathbf{q}}^{\mathrm{R}}_{q,h}Z_{\mathrm{R}}^\top=\mathbf{q}^{\mathrm{R}}_{q,h}(\widehat{K}^{\mathrm{R}})^{\top}$:
attention reads exactly the dequantized cache of Sections~\ref{sec:method-latent}
and~\ref{sec:method-kR} without ever reconstructing it. Three small products
remain at runtime, all acting on the current token and hence independent of
context length: one $d_{\mathrm{c}}\times d_{\mathrm{c}}$ on the latent write path, one
$d^{\mathrm{R}}\times d^{\mathrm{R}}$ on the positional-key write path, and one
$d^{\mathrm{R}}\times d^{\mathrm{R}}$ per head on the query path.

\section{Experiments}
\label{sec:results}

\subsection{Setup}
\label{sec:setup}

\paragraph{Models and benchmarks.}
We evaluate four MLA models: GLM-4.7-Flash \citep{glm45}, GLM-5.2-FP8
\citep{glm5}, Moonlight-16B-A3B-Instruct \citep{moonlight}, and
DeepSeek-Coder-V2-Lite-Instruct \citep{deepseekcoderv2}, all with $d_{\mathrm{c}}=512$
and $d^{\mathrm{R}}=64$. Benchmarks cover reasoning and coding on GPQA-Diamond
\citep{gpqa}, AIME25 \citep{matharena}, MATH-500 \citep{math500}, GSM8K \citep{gsm8k}, and
HumanEval \citep{humaneval}, with ARC-Challenge \citep{arc} and PIQA
\citep{piqa} in place of AIME25 and GPQA-Diamond on Moonlight and DS-V2-Lite,
and long-context retrieval on the eight RULER NIAH subtasks \citep{ruler} from
$4$K to $128$K. More details are given in Appendix~\ref{app:protocol}.

\paragraph{Baselines.}
Among existing KV-cache quantization methods, KVarN \citep{kvarn} reports MLA
experiments, compressing the latent with a Hadamard rotation and a dual-scaling
variance normalization while leaving the decoupled positional key in full
precision. We also implement two standard KV-cache transform methods, QuaRot
\citep{quarot} and OSCAR \citep{oscar}, in the same setting: transform the
latent cache, quantize it to low bits, and leave the decoupled positional key
in full precision.

\paragraph{Implementation.}
The transforms are fitted on $128$ RedPajama \citep{redpajama} sequences of $64$K tokens. The
latent uses asymmetric per-token INT2 with group size $128$ and a clipping
ratio of $0.94$, and $\mathbf{k}^{\mathrm{R}}$ uses asymmetric per-channel INT4 within
each cache page. Following KVarN, a cache page is quantized only once it is
full, and the open page stays in BF16 for all quantized methods. We report
effective bits per cached element (BPE), which include the FP16 scale and
minimum; the full cache costs $2.47$ BPE, or $2.50$ on GLM-5.2, whose
$64$-token pages double the metadata of $\mathbf{k}^{\mathrm{R}}$.

\subsection{Main results}
\label{sec:results-short}

\begin{table}[t]
\centering
\caption{\textbf{Reasoning and coding tasks.} \textbf{BPE} is effective bits per
cached element, and $\Delta$ is the mean change from BF16 over the five tasks.
}
\label{tab:glm}
\footnotesize
\setlength{\tabcolsep}{2pt}
\setlength{\dashlinedash}{2pt}
\setlength{\dashlinegap}{2pt}
\begin{tabular}{llccccccc}
\toprule
Model & Method & BPE & AIME25 & GPQA & MATH-500 & GSM8K & HumanE & $\Delta$ \\
\midrule
 & BF16                 & 16.00 & 93.33 & 77.27 & 97.40 & 95.22 & 93.29 & --- \\
 & KVarN INT3           & 4.67  & 90.00 & 73.23 & \textbf{98.20} & 95.60 & 92.07 & $-1.48$ \\
 & QuaRot INT2          & 3.78  & 86.67 & 69.19 & \textbf{98.20} & 94.09 & 86.59 & $-4.35$ \\
 & OSCAR INT2           & 3.78  & 87.78 & 72.06 & 97.40 & 95.75 & 94.31 & $-1.84$ \\
\ours
\multirow{-5}{*}{GLM-4.7}
 & \textbf{RAQ-MLA} & \textbf{2.47} & \textbf{92.22} & \textbf{75.25} & 97.67 & \textbf{96.08} & \textbf{94.92} & $\mathbf{-0.08}$ \\
\cmidrule(lr){1-9}
 & BF16                 & 16.00 & 90.00 & 86.87 & 97.80 & 97.50 & 99.39 & --- \\
 & QuaRot INT2        & 3.78  & 86.67 & 78.28 & 98.20 & 97.80 & 96.34 & $-2.85$ \\
 & OSCAR INT2           & 3.78  & \textbf{93.33} & \textbf{85.86} & \textbf{98.60} & 97.62 & 98.58 & $\mathbf{+0.49}$ \\
\ours
\multirow{-4}{*}{GLM-5.2}
 & \textbf{RAQ-MLA} & \textbf{2.50} & 92.22 & 84.85 & 97.80 & \textbf{97.90} & \textbf{98.78} & $0.00$ \\
\addlinespace[0.4ex]
\hdashline
\addlinespace[0.65ex]
 & & & ARC-C & PIQA & MATH-500 & GSM8K & HumanE & $\Delta$ \\
\midrule
 & BF16                 & 16.00 & 52.22 & 52.56 & 63.20 & 85.14 & 64.02 & --- \\
 & QuaRot INT2          & 3.78  & 45.14 & 33.13 & 32.20 & 59.29 & 44.51 & $-20.57$ \\
 & OSCAR INT2           & 3.78  & 50.77 & 48.04 & 57.13 & 80.42 & \textbf{69.21} & $-2.31$ \\
\ours
\multirow{-4}{*}{Moonlight}
 & \textbf{RAQ-MLA} & \textbf{2.47} & \textbf{51.79} & \textbf{52.67} & \textbf{61.80} & \textbf{83.85} & 67.27 & $\mathbf{+0.05}$ \\
\cmidrule(lr){1-9}
 & BF16                 & 16.00 & 69.97 & 75.30 & 58.00 & 84.91 & 78.05 & --- \\
 & QuaRot INT2          & 3.78  & 60.67 & 68.39 & 51.80 & 75.28 & 70.73 & $-7.87$ \\
 & OSCAR INT2           & 3.78  & \textbf{68.69} & \textbf{74.54} & \textbf{61.13} & 82.74 & \textbf{79.68} & $\mathbf{+0.11}$ \\
\ours
\multirow{-4}{*}{DS-V2-Lite}
 & \textbf{RAQ-MLA} & \textbf{2.47} & 68.60 & \textbf{74.54} & 60.60 & \textbf{83.52} & 78.25 & $-0.14$ \\
\bottomrule
\end{tabular}
\end{table}

\begin{table}[t]
\centering
\caption{\textbf{RULER-NIAH on GLM-4.7-Flash.} Each cell averages eight
subtasks of $100$ samples.}
\label{tab:ruler}
\begin{tabular*}{\textwidth}{@{\extracolsep{\fill}}lcccccccc}
\toprule
Method & BPE & 4K & 8K & 16K & 32K & 64K & 128K & Avg. \\
\midrule
BF16            & 16.00 & 100.00 & 99.75 & 99.31 & 98.44 & 96.97 & 87.22 & 96.95 \\
OSCAR           & 3.78  & \textbf{99.56} & 98.64 & \textbf{98.34} & \textbf{96.01} & 90.39 & 78.65 & 93.60 \\
\ours \textbf{RAQ-MLA} & \textbf{2.66} & 99.38 & \textbf{99.06} & 97.96 & 95.74 & \textbf{90.69} & \textbf{79.13} & \textbf{93.66} \\
\bottomrule
\end{tabular*}
\end{table}

\paragraph{Reasoning and coding.}
Table~\ref{tab:glm} shows that RAQ-MLA keeps BF16 accuracy at the lowest bit
width of all methods. At about $2.5$ bits, it stays within $0.5$ points of BF16
on all four models ($-0.08$, $0.00$, $+0.05$, $-0.14$), whereas QuaRot and OSCAR
store $3.78$ BPE and KVarN $4.67$ (Figure~\ref{fig:capacity-accuracy}). On
GLM-4.7, every baseline loses more despite its larger cache: QuaRot $4.35$
points, OSCAR $1.84$ and KVarN $1.48$; QuaRot also falls by $20.57$ points on
Moonlight and $7.87$ on DS-V2-Lite. OSCAR, the strongest baseline, is slightly
ahead on GLM-5.2 and DS-V2-Lite ($+0.49$ and $+0.11$) but loses $2.31$ points on
Moonlight, and it needs about $1.5\times$ the cache of RAQ-MLA.

\paragraph{Long-context retrieval.}
Table~\ref{tab:ruler} reports RULER NIAH on GLM-4.7-Flash. Here RAQ-MLA keeps
a wider quantizer at the first and last two layers (an INT4 latent and an INT8
positional key), for $2.66$ BPE on average. It matches OSCAR on average
($93.66$ against $93.60$) with a $29.6\%$ smaller cache ($2.66$ against $3.78$
BPE), and the two stay within $0.5$ points of each other at every length. Both
trail a BF16 control that itself falls from $96.97$ at $64$K to $87.22$ at
$128$K.

\subsection{Ablation study}
\label{sec:results-ablation-kR}

\begin{figure}[t]
\begin{minipage}[c]{0.52\textwidth}
\centering
\captionsetup{skip=2pt}
\captionof{table}{\textbf{Component ablation on RAQ-MLA}, GLM-4.7-Flash.}
\label{tab:component-ladder}
\footnotesize
\setlength{\tabcolsep}{2pt}
\begin{tabular}{cl@{\hspace{7pt}}l@{\hspace{7pt}}crr}
\toprule
Step & Latent & $\mathbf{k}^{\mathrm{R}}$ & BPE & AIME25 & GPQA \\
\midrule
A & naive & BF16 & 3.78 & 83.33 & 71.72 \\
B & clip + tail & BF16 & 3.78 & 86.67 & 76.26 \\
C & Hadamard & BF16 & 3.78 & 86.67 & 69.19 \\
D & OSCAR & BF16 & 3.78 & 87.78 & 72.06 \\
E & value-aware & BF16 & 3.78 & 92.22 & 76.09 \\
F & value-aware & direct & 2.47 & 87.78 & 74.24 \\
\ours G & value-aware & score-aware & 2.47 & 92.22 & 75.25 \\
\bottomrule
\end{tabular}
\end{minipage}\hfill
\begin{minipage}[c]{0.45\textwidth}
\centering
\captionsetup{justification=raggedright,singlelinecheck=false,skip=2pt}
\includegraphics[width=\linewidth]{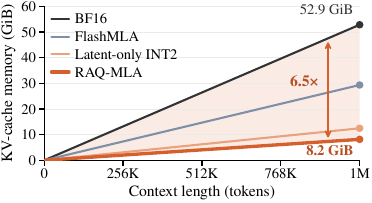}
\caption{\textbf{Theoretical KV-cache memory} of one GLM-4.7-Flash
sequence.}
\label{fig:kv-memory}
\end{minipage}
\end{figure}

\paragraph{Component ablation.}
Table~\ref{tab:component-ladder} adds the components of RAQ-MLA one at a time.
Following OSCAR and KVarN, B adds clipping and keeps the open cache page in
BF16. On top of B, C--E compare three latent transforms, and our value-aware
transform is the most accurate on both tasks: $92.22$ on AIME25 and $76.09$ on
GPQA, against $87.78$ and $72.06$ for OSCAR and $86.67$ and $69.19$ for
Hadamard. When $\mathbf{k}^{\mathrm{R}}$ is quantized as well, our score-aware transform
(G) improves over direct INT4 (F) from $87.78$ to $92.22$ on AIME25 and from
$74.24$ to $75.25$ on GPQA at the same $2.47$ BPE.

\subsection{Efficiency}
\label{sec:results-efficiency}

\begin{wraptable}{r}{0.53\textwidth}
\vspace{-\intextsep}
\centering
\caption{\textbf{Kernel latency and TPOT on H20 GPUs.}}
\label{tab:packed-runtime-latency}
\small
\setlength{\tabcolsep}{10.5pt}
\begin{tabular}{lrr}
\toprule
Method & Kernel ($\mu$s/layer) & TPOT (ms) \\
\midrule
FlashMLA              & 17.4          & 6.56 \\
\ours \textbf{RAQ-MLA} & \textbf{12.6} & \textbf{5.97} \\
\midrule
Speedup               & \textbf{1.38$\times$} & \textbf{1.10$\times$} \\
\bottomrule
\end{tabular}
\end{wraptable}
RAQ-MLA reduces the KV cache by $6.5\times$ compared with BF16 and by
$3.6\times$ compared with the FP8 cache of FlashMLA \citep{flashmla}
(Figure~\ref{fig:kv-memory}). On two H20 GPUs with TP2 and batch size $1$, the
decode kernel at $512$ context drops from $17.4$ to $12.6$~$\mu$s per layer
against FlashMLA ($1.38\times$), and the time per output token at $512$ input
and $512$ output tokens from $6.56$ to $5.97$~ms ($1.10\times$), because each
decode step reads less cache (Table~\ref{tab:packed-runtime-latency}).
TPOT stays $6.8$--$9.0\%$ lower for outputs of $1$K to $8$K tokens, and
applying $A_{\mathrm{c}}$ inside the cache-write kernel costs $1.7$--$2.4$~$\mu$s
per layer, under $2\%$ of TPOT at batch size $1$
(Appendix~\ref{app:full-results}).

\section{Conclusion}
\label{sec:conclusion}

We showed that the two tensors of the MLA cache play different roles in
attention: the decoupled positional key dominates the attention score, whereas
error in the latent reaches the output mainly through the value readout.
RAQ-MLA therefore stores each tensor in an invertible basis fitted, without
training, against the error it delivers to the attention output. It compresses
the full MLA cache, positional key included, to about $2.5$ bits per element, a
$6.5\times$ capacity gain over BF16, within $0.5$ points of BF16 on reasoning
and coding across the GLM, DeepSeek, and Moonlight families.

\subsection*{AI use statement}
We used a generative AI assistant (Claude by Anthropic, accessed through the
Cursor editor) only for auxiliary tasks: polishing the wording and checking the
grammar of text written by the authors; formatting the LaTeX source, tables,
and references, and checking them against the submission requirements; and
restyling figures (fonts, tick labels, and layout) through the plotting
scripts, without changing the plotted data. The core content of this work is
the authors' own: the research questions, the analysis of the MLA cache, the
RAQ-MLA method and its derivations, the design, implementation, and execution
of all experiments, and the interpretation of the results. The authors reviewed
every AI-assisted edit and verified that no figure's data were altered. We take
full responsibility for the entire content of this paper, including any text,
figures, or code produced with the aid of generative AI.

\subsection*{Reproducibility statement}
Section~\ref{sec:method} specifies RAQ-MLA: the objective in
\eqref{eq:general-objective}, the role metrics of the latent and the positional
key (Sections~\ref{sec:method-latent} and~\ref{sec:method-kR}), the alternating
fit (Section~\ref{sec:method-fit}), and how the fitted transforms are folded
into inference (Section~\ref{sec:method-serving}). Appendix~\ref{app:algorithm}
gives pseudocode for fitting and decoding
(Algorithms~\ref{alg:raq-mla-fit} and~\ref{alg:raq-mla-decode}) and the proof
of Theorem~\ref{thm:descent}. Section~\ref{sec:setup} lists the evaluated
models, benchmarks, baselines, and calibration data.
Appendix~\ref{app:quantizer-accounting} gives the quantizer settings and the
accounting of effective bits per element (Table~\ref{tab:app-quantizer}), and
Appendix~\ref{app:evaluation-protocol} gives the serving stack, evaluation
toolkits, and sampling configuration (Table~\ref{tab:app-serving}). The
hardware and software settings of the latency measurements are stated with each
result in Section~\ref{sec:results-efficiency} and
Appendix~\ref{app:full-results}.

\bibliography{refs}
\bibliographystyle{iclr2027_conference}

\appendix
\section{Related Work}
\label{sec:related}

\paragraph{KV-cache quantization.}
Post-training quantization of weights and activations is well established
\citep{gptq,awq,smoothquant}. As context grows, the KV cache rather than the
weights dominates decoding memory, and the same tools have been carried over to
the cache. Existing KV-cache quantizers choose their granularity from the
statistics of the stored tensor: keys have persistent channel-wise outliers and
values do not, so KIVI \citep{kivi} quantizes keys per channel and values per
token, keeping a short window of recent tokens in full precision. KVQuant
\citep{kvquant} quantizes keys per channel before RoPE, fits non-uniform
datatypes, and stores a small fraction of outlier values in a sparse
full-precision matrix. At two bits, SKVQ \citep{skvq} reorders channels and
clips dynamically, with a full-precision sliding window of recent tokens. Finer
groups are not free, since each group stores its own scale and zero point, so
we report effective bits per cached element throughout. All of these methods
choose a tensor's quantization policy from that tensor's own statistics, not
from how attention reads it.

\paragraph{Transforms for quantization.}
Rotating the cache before rounding is the most common remedy for outliers at low
bit widths. QuaRot \citep{quarot} applies Hadamard rotations that leave the
network output unchanged and quantizes weights, activations, and the KV cache to
four bits; SpinQuant \citep{spinquant} learns these rotations instead. RotateKV
\citep{rotatekv} rotates keys before RoPE with a grouped-head Hadamard transform
and outlier-aware channel reordering. OSCAR \citep{oscar} derives fixed
rotations and clipping thresholds offline from attention-aware covariance
statistics, combining their eigenbases with a Hadamard transform. OScaR
\citep{oscar_meituan}, a different method despite the similar name, targets
token-norm imbalance with canalized rotation and omni-token scaling. KVarN
\citep{kvarn} applies a Hadamard rotation followed by variance normalization
along both axes of the key and value matrices, without calibration. Beyond
orthogonal transforms, AffineQuant \citep{affinequant} and FlatQuant
\citep{flatquant} learn general invertible affine transforms for low-bit LLM
quantization by gradient descent on layer or block outputs.

\paragraph{MLA cache quantization.}
Few methods target the MLA cache directly. SnapMLA \citep{snapmla} quantizes the
latent to FP8 per token and keeps the positional key in BF16, after observing
that FP8 raises the error of the RoPE part by an order of magnitude. The FP8
cache format of FlashMLA \citep{flashmla} likewise stores the latent in FP8 and
the positional key in BF16. The official vLLM implementation of KVarN
\citep{kvarn} supports MLA models by quantizing only the latent, to INT4. All of
them leave $\mathbf{k}^{\mathrm{R}}$ in full precision.

\paragraph{Distinction from existing methods.}
RAQ-MLA is a low-bit post-training quantization method designed for the MLA
cache. The general KV-cache quantizers and transforms above are designed around
the keys and values of standard attention, and the MLA cache quantizers keep the
latent in FP8 or INT4 and leave $\mathbf{k}^{\mathrm{R}}$ in BF16; even with a 2-bit
latent, a BF16 positional key keeps the cache at $3.78$ effective bits per
element or more. RAQ-MLA quantizes the full MLA cache, including the positional
key, to about $2.5$ bits per cached element. It treats the two cached tensors
according to their roles in attention: the latent is quantized per token with a
value-aware transform, and the positional key per channel with a score-aware
transform. Hadamard is fixed, and KVarN and OSCAR derive their transforms from
data statistics before any rounding takes place; RAQ-MLA instead fits general
invertible transforms to the error that rounding leaves at the attention output.
AffineQuant and FlatQuant also learn invertible transforms, but for weights and
activations and by backpropagating through the network, whereas RAQ-MLA is
fitted by alternating minimization without any gradient through the model.

\section{Overall Algorithm}
\label{app:algorithm}

This appendix states the two halves of RAQ-MLA in full: the offline fit of
Section~\ref{sec:method-fit}, which produces one transform per layer and per
cached tensor, and the decode-time computation of
Section~\ref{sec:method-serving}, which consumes it.

\paragraph{Fitting.}
Algorithm~\ref{alg:raq-mla-fit} runs once per layer and per cached tensor and
does not train the model. The unquantized model is run once on the calibration
and validation splits to collect the cached tensor together with the
statistics its role metric needs: the exposure weights $\gamma_{t,h}$ for the
latent and the positional queries $\mathbf{q}^{\mathrm{R}}$ for the positional key.
$M_t$ is never formed explicitly; the objective and the inner solve use these
factors directly. Each round freezes the quantized tensor, solves the proximal
problem of \eqref{eq:transform-update} by conjugate gradient, and line-searches
toward the solution. Every trial step $B=A^{-1}+\alpha(\tilde{B}-A^{-1})$ is
requantized and accepted only if the objective decreases,
$F(B^{-1})<F(A)$, and $\mathrm{cond}(B)\le\kappa_{\max}$; if no step is
accepted, $\lambda$ is increased tenfold, at most twice, before the fit
stops. We run the fit for each $\lambda$ in a small grid and return, among the
initialization and all accepted iterates, the transform with the lowest
attention-output error on the validation split.

\begin{algorithm}[!htb]
\caption{Fitting one RAQ-MLA transform.}
\label{alg:raq-mla-fit}
\begin{algorithmic}[1]
\Require calibration and validation splits; quantizer $\mathcal{Q}$; initial
         transform $A^{(0)}$; proximal weights $\Lambda$; round budget $R$;
         condition bound $\kappa_{\max}$
\Ensure transform $A$
\State Run the unquantized model on both splits; collect $X$ and the factors of
       $M_t$ ($\gamma_{t,h}$ and $\mathbf{W}^{\mathrm{UV}}$ for the latent, $\mathbf{q}^{\mathrm{R}}$ for
       the positional key)
\State $\mathcal{C}\leftarrow\{A^{(0)}\}$
\For{$\lambda\in\Lambda$}
  \State $A\leftarrow A^{(0)}$
  \For{$r=1,\dots,R$}
    \State $Z\leftarrow\mathcal{Q}(XA)$
    \State $\tilde{B}\leftarrow\arg\min_B\ \mathcal{D}(ZB)
           +\lambda\|B-A^{-1}\|_F^2$
           \Comment{strongly convex; solved by CG}
    \State Find the largest accepted $\alpha\in\{1,\tfrac12,\dots\}$
           \Comment{requantized line search}
    \State \textbf{if} no $\alpha$ is accepted \textbf{then break}
    \State $A\leftarrow\bigl(A^{-1}+\alpha(\tilde{B}-A^{-1})\bigr)^{-1}$;\ \
           $\mathcal{C}\leftarrow\mathcal{C}\cup\{A\}$
  \EndFor
\EndFor
\State \Return the transform in $\mathcal{C}$ with the lowest attention-output
       error on the validation split
\end{algorithmic}
\end{algorithm}

\paragraph{Inference.}
The fitted transforms enter decoding in two ways. The inverse of the latent
transform is folded once, before serving, into the up-projections by
\eqref{eq:latent-folding}, so attention reads the cached latent directly in its
learned basis. The forward transforms cannot be folded, because RMSNorm
precedes $A_{\mathrm{c}}$ and RoPE precedes $A_{\mathrm{R}}$; each is applied to the current token on
the write path. Likewise, $A_{\mathrm{R}}^{-\top}$ must follow RoPE on the query side,
which is admissible because $\mathbf{q}^{\mathrm{R}}$ is never cached.
Algorithm~\ref{alg:raq-mla-decode} gives the resulting decode step, written
with row vectors to match Section~\ref{sec:method}.

\begin{algorithm}[t]
\caption{One RAQ-MLA decode step at a layer.}
\label{alg:raq-mla-decode}
\begin{algorithmic}[1]
\Require hidden state $\mathbf{h}$; transforms $A_{\mathrm{c}}$, $A_{\mathrm{R}}$; folded
         up-projections $(\widetilde{\mathbf{W}}_{h}^{\mathrm{UK}})^{\top}=A_{\mathrm{c}}^{-1}(\mathbf{W}_{h}^{\mathrm{UK}})^{\top}$ and
         $(\widetilde{\mathbf{W}}_{h}^{\mathrm{UV}})^{\top}=A_{\mathrm{c}}^{-1}(\mathbf{W}_{h}^{\mathrm{UV}})^{\top}$; caches $Z_{\mathrm{c}}$, $Z_{\mathrm{R}}$
         (quantized pages followed by an open BF16 page)
\Statex \textbf{Write path}
\State $\widetilde{\mathbf{c}}^{\mathrm{KV}}\leftarrow
       \mathrm{RMSNorm}\!\left(\mathbf{h}(\mathbf{W}^{\mathrm{DKV}})^{\top}\right)A_{\mathrm{c}}$
       \Comment{$d_{\mathrm{c}}\times d_{\mathrm{c}}$ product}
\State $\widetilde{\mathbf{k}}^{\mathrm{R}}\leftarrow
       \mathrm{RoPE}\!\left(\mathbf{h}(\mathbf{W}^{\mathrm{KR}})^{\top}\right)A_{\mathrm{R}}$
       \Comment{$d^{\mathrm{R}}\times d^{\mathrm{R}}$ product}
\State Append $\widetilde{\mathbf{c}}^{\mathrm{KV}}$ and $\widetilde{\mathbf{k}}^{\mathrm{R}}$ to
       the open page; once the page is full, replace it by its quantized form
\Statex \textbf{Read path, per head $h$}
\State $\mathbf{q}^{\mathrm{C}}_h,\ \mathbf{q}^{\mathrm{R}}_h\leftarrow$ query projections of
       $\mathbf{h}$, with RoPE applied to $\mathbf{q}^{\mathrm{R}}_h$
\State $\widetilde{\mathbf{q}}^{\mathrm{R}}_h\leftarrow\mathbf{q}^{\mathrm{R}}_hA_{\mathrm{R}}^{-\top}$
       \Comment{$d^{\mathrm{R}}\times d^{\mathrm{R}}$ product; cancels $A_{\mathrm{R}}$}
\State $L_h\leftarrow\left(
       \mathbf{q}^{\mathrm{C}}_h\widetilde{\mathbf{W}}_{h}^{\mathrm{UK}}Z_{\mathrm{c}}^\top
       +\widetilde{\mathbf{q}}^{\mathrm{R}}_hZ_{\mathrm{R}}^\top\right)\!
       \big/\sqrt{d_{\mathrm{h}}+d^{\mathrm{R}}}$
\State $\mathbf{o}_h\leftarrow\mathrm{softmax}\!\left(L_h\right)
       Z_{\mathrm{c}}(\widetilde{\mathbf{W}}_{h}^{\mathrm{UV}})^{\top}$
\State \Return the head outputs combined through $\mathbf{W}^{\mathrm{O}}$
\end{algorithmic}
\end{algorithm}

Relative to an unmodified MLA implementation, RAQ-MLA adds three small matrix
products per decode step, all acting on the current token and hence independent
of context length: one $d_{\mathrm{c}}\times d_{\mathrm{c}}$ product on the latent write path
(line~1), one $d^{\mathrm{R}}\times d^{\mathrm{R}}$ product on the positional-key write path
(line~2), and one $d^{\mathrm{R}}\times d^{\mathrm{R}}$ product per head on the query path
(line~5). Attention itself runs on the quantized cache with the same structure
as unmodified MLA.

\paragraph{Proof of Theorem~\ref{thm:descent}.}
Let $G_r(B)=\mathcal{D}(Z^{(r)}B)+\lambda\|B-B^{(r)}\|_F^2$, whose minimizer
$\tilde{B}$ is the update of \eqref{eq:transform-update}. Because
$G_r(B^{(r)})=F(A^{(r)})$ and $G_r(\tilde{B})\le G_r(B^{(r)})$, the choice
$B^{(r+1)}=\tilde{B}$ gives
$\mathcal{D}(Z^{(r)}B^{(r+1)})\le F(A^{(r)})-\lambda\Delta_r$, and the
hypothesis of the theorem replaces the left-hand side by $F(A^{(r+1)})$. The
same bound holds for the damped step $B^{(r+1)}=B^{(r)}+\alpha(\tilde{B}-B^{(r)})$
accepted by Algorithm~\ref{alg:raq-mla-fit}, with $\lambda$ the value used in
that round, since $G_r$ is convex and does not exceed $F(A^{(r)})$ at either end
of the segment. The sequence $F(A^{(r)})$ is therefore nonincreasing and
bounded below by $0$, so it converges, and summing the descent inequality over
$r$ gives $\sum_r\lambda\Delta_r\le F(A^{(0)})$, hence $\Delta_r\to0$. Fitting
therefore cannot degrade its initialization on the calibration objective under
the stated condition; Algorithm~\ref{alg:raq-mla-fit} also enforces descent
directly, since it accepts a step only if the requantized objective decreases.
Because the rounding in $\mathcal{Q}$ makes $F$ discontinuous in $A$, we claim
no stationarity or optimality; how the fitted transforms transfer to unseen
data is evaluated in Section~\ref{sec:results}.

\section{Analysis of the RAQ-MLA Fit}
\label{app:more-ablations}

\begin{figure}[t]
\centering
\includegraphics[width=\linewidth]{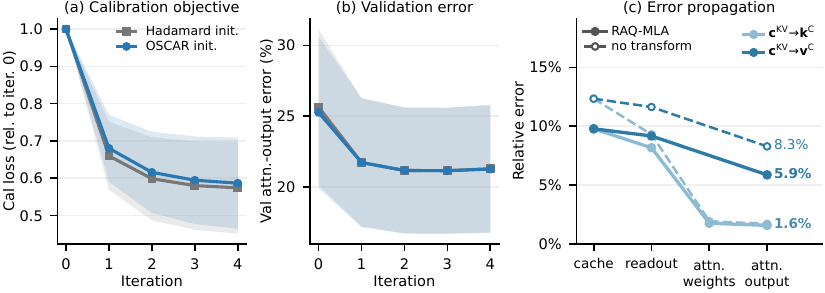}
\caption{\textbf{Analysis of the RAQ-MLA latent fit on GLM-4.7-Flash.}
(a, b) Calibration objective relative to iteration~$0$ and Val
attention-output error per round, mean$\pm$std over the $47$ INT2 layers along
each layer's Val-selected damping trajectory, from Hadamard and OSCAR
initializations. (c) Relative error of the latent at each stage under the INT4
quantizer of Figure~\ref{fig:error-fate}, with the fitted $A_{\mathrm{c}}$
(solid) and without a transform (dashed).}
\label{fig:app-raqmla-analysis}
\end{figure}

Figure~\ref{fig:app-raqmla-analysis} examines the latent fit on GLM-4.7-Flash
from three angles. First, the fit converges within a few rounds (a, b).
Averaged over the $47$ INT2 layers, one update lowers the calibration
objective to about two thirds of its initial value and the Val
attention-output error from $25.3\%$ to $21.7\%$. The Val error levels off at
about $21.1\%$ from the second round. Hadamard and OSCAR initializations follow
nearly the same curves. Second, the choice of role metric matters: minimizing
the latent's own quantization error, i.e., fitting with $M_t=I$, does not
minimize the error that reaches the attention output. Under the same
initialization, optimizer, and budget, the $M_t=I$ fit lowers the Val
attention-output error from $25.3\%$ to $23.3\%$, whereas the role metric
lowers it to $21.1\%$ and is better on all $47$ layers. Third, the fitted
transform reduces the error on both latent paths (c). Under the INT4 quantizer
of Figure~\ref{fig:error-fate}, $A_{\mathrm{c}}$ lowers the error that reaches
the attention output through the value from $8.3\%$ to $5.9\%$, and through the
content key from $1.7\%$ to $1.6\%$, even though the value-aware objective does
not model the content-key path.

\section{Additional Diagnostic Figures}
\label{app:more-visualizations}

\paragraph{Logit attribution.}
\label{app:logit-attribution}
The logit attribution in Section~\ref{sec:motivation} is the share of the
squared logit norm carried by each group of channels. For head $h$, query
position $q$, and cached token $t$, the content contribution of a channel set
$\mathcal{S}$ is
$(\mathbf{q}^{\mathrm{C}}_{q,h}\mathbf{W}^{\mathrm{UK}}_{h})_{\mathcal{S}}
(\mathbf{c}^{\mathrm{KV}}_{t})_{\mathcal{S}}^{\top}$,
and the positional contribution is
$(\mathbf{q}^{\mathrm{R}}_{q,h})_{\mathcal{S}}(\mathbf{k}^{\mathrm{R}}_{t})_{\mathcal{S}}^{\top}$.
Outlier channels are the $10\%$ of largest magnitude. Stacking each group's
contributions over heads, queries, and tokens $t\le q$ gives $L_g$, with
$L=\sum_g L_g$. The share is
$a_g=\langle L_g, L\rangle_F\big/\|L\|_F^2$,
pooled by summing the numerator and the denominator over all $47$ layers. The
reported $92.5\%$ is the positional key's share.

\begin{figure}[t]
\centering
\includegraphics[width=\textwidth]{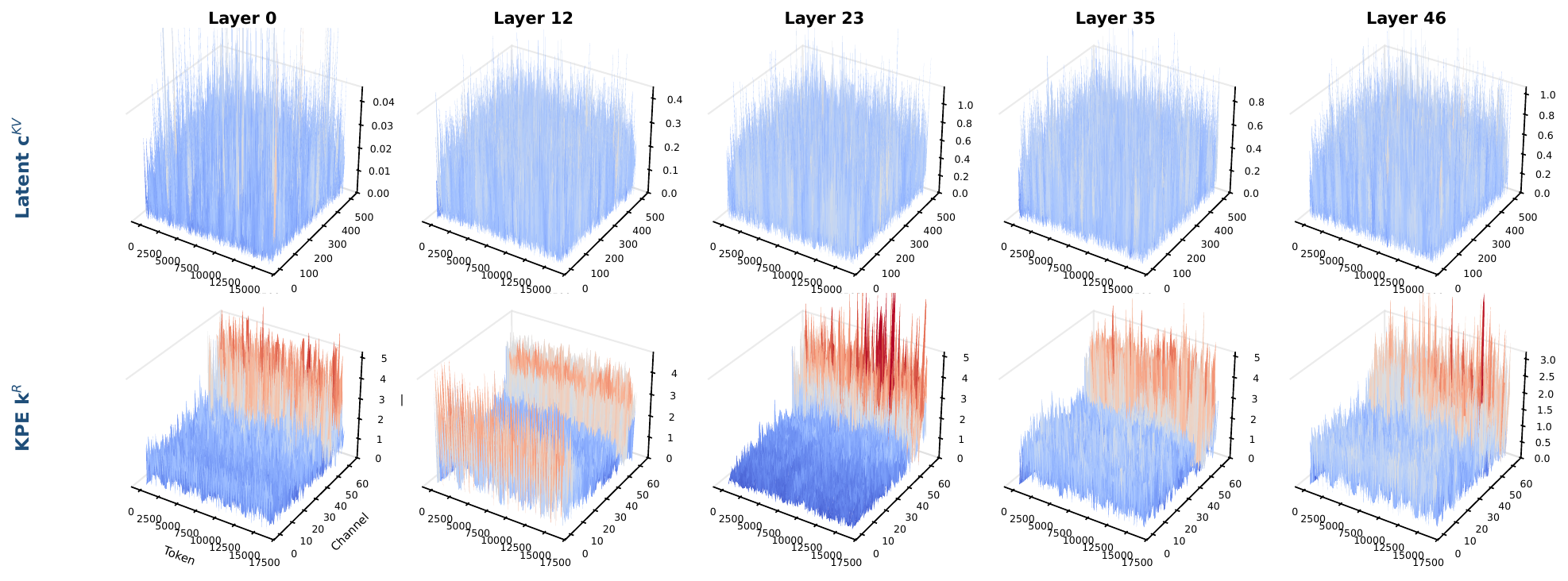}
\caption{\textbf{The latent and positional key retain different channel
structures throughout depth.} Token--channel magnitude surfaces for the
normalized compressed latent $\mathbf{c}^{\mathrm{KV}}$ (top) and post-RoPE positional
key $\mathbf{k}^{\mathrm{R}}$ (bottom) at five approximately uniformly spaced layers of
GLM-4.7-Flash. Each panel uses its own $99$th-percentile vertical scale, so the
comparison concerns structure rather than absolute magnitude. The latent
remains diffuse across channels, whereas the positional key exhibits persistent
fixed-channel bands from early to late layers.}
\label{fig:app-outlier-depth}
\end{figure}

\paragraph{Outlier structure across layer depth (extension of
Figure~\ref{fig:outlier-dist}).}
Figure~\ref{fig:outlier-dist} contrasts the token--channel magnitude patterns
of the compressed latent and positional key at one representative layer. The
depth extension in Figure~\ref{fig:app-outlier-depth} shows the same two cached
tensors at five approximately uniformly spaced layers.

\begin{figure}[t]
\centering
\begin{minipage}[b]{0.585\textwidth}
\centering
\includegraphics[width=\linewidth]{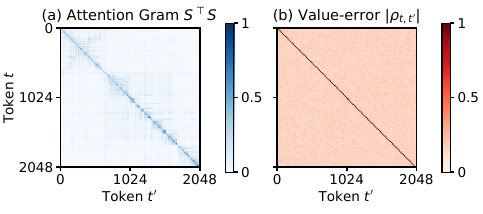}
\end{minipage}\hfill
\begin{minipage}[b]{0.395\textwidth}
\centering
\includegraphics[width=\linewidth]{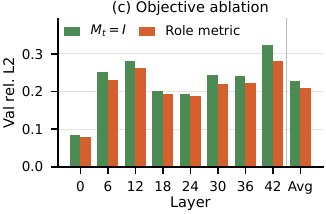}
\end{minipage}
\caption{\textbf{Value-readout errors are nearly uncorrelated across tokens,
although attention reads nearby tokens together.} (a--b) Layer~$23$ of
GLM-4.7-Flash, first $2048$ tokens of a calibration prompt, deployed INT2
quantizer with the fitted RAQ-MLA transform, visualized at $16$-token
resolution. (a) Attention Gram $S^\top S$, normalized by its diagonal and
averaged over heads. (b) Magnitude of the correlation $\rho_{t,t'}$ between
the value-readout errors at tokens $t$ and $t'$, pooled over heads. (c) Val
attention-output error of the latent transform fitted with $M_t=I$ or with the
role metric under the same fitting pipeline.}
\label{fig:app-value-crosstoken}
\end{figure}

\paragraph{Token independence of the value-path error.}
The value-aware objective of Section~\ref{sec:method-latent} keeps only the
diagonal of $S_h^\top S_h$, which assumes that quantization errors at different
tokens are nearly uncorrelated. Figure~\ref{fig:app-value-crosstoken} tests
this assumption at layer~$23$ over the first $2048$ tokens of a calibration
prompt, under the deployed INT2 quantizer and the fitted RAQ-MLA transform. The
attention Gram is far from diagonal: nearby tokens are read together by the
same queries. The value-readout errors, in contrast, are correlated only with
themselves; the mean off-diagonal $|\rho_{t,t'}|$ is $0.044$. Summed over all
token pairs, the residual off-diagonal terms still account for $14.2\%$ of the
value-path distortion at this layer ($13.1\%$ for Hadamard and $16.3\%$
without a transform). Because this share is similar across transforms,
dropping them mainly rescales $\mathcal{D}_{\mathrm{c}}$ rather than reordering candidate
transforms (Figure~\ref{fig:role-metric-correlation}).

\begin{figure}[t]
\centering
\includegraphics[width=\textwidth]{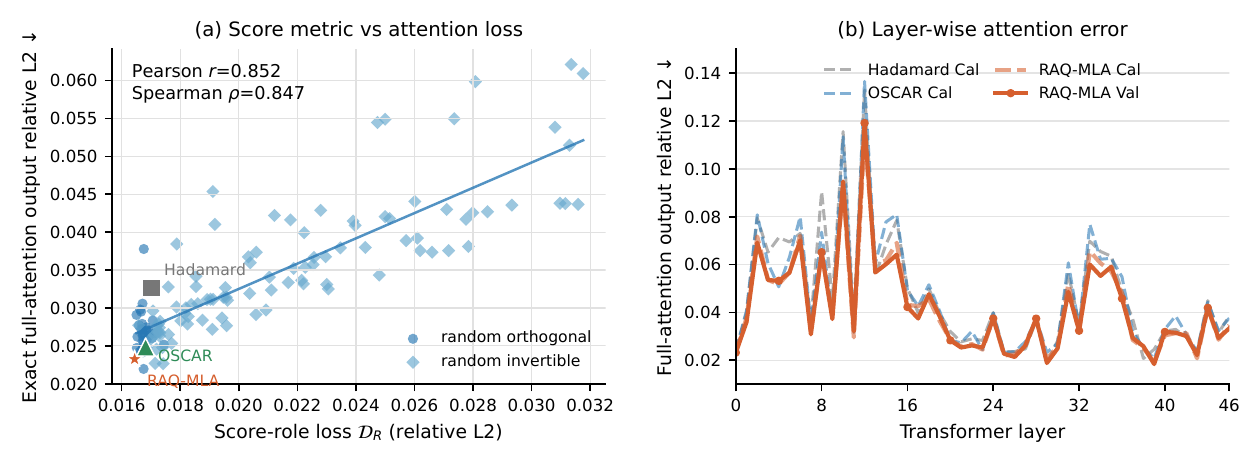}
\caption{\textbf{Evaluation of the positional-key role metric.}
(a) Score-role loss $\mathcal{D}_{\mathrm{R}}$ against the exact attention-output error at
layer~$23$ across $128$ random transforms; Hadamard, OSCAR, and RAQ-MLA are
excluded from the correlations. (b) Layer-wise attention-output error; baselines
are shown on calibration prompts only.}
\label{fig:app-kpe-role-metric}
\end{figure}

\paragraph{Positional-key role metric (extension of
Figure~\ref{fig:role-metric-validation}).}
Figure~\ref{fig:role-metric-validation} validates the role metric of the
latent; Figure~\ref{fig:app-kpe-role-metric} repeats both tests for the
score-aware metric $M^{\mathrm{R}}_t$ of Section~\ref{sec:method-kR}. Only the positional
key is quantized, to INT4 per channel over each cache page, while the latent
stays in BF16. At layer~$23$, we score $32$ random orthogonal and $96$ random
invertible $64\times64$ transforms by both $\mathcal{D}_{\mathrm{R}}$ and the exact
attention-output error. The two remain strongly correlated (Pearson $r=0.852$,
Spearman $\rho=0.847$), though less tightly than for the latent ($r=0.992$).
Across layers, fitting $A_{\mathrm{R}}$ lowers the mean attention-output error on
validation prompts from $0.050$ for Hadamard and $0.047$ for OSCAR to $0.042$,
and is below OSCAR at $43$ of the $47$ layers and below Hadamard at $40$. Its
calibration and validation curves nearly coincide, differing by $3.7\%$ on
average.

\section{Additional Implementation and Evaluation Details}
\label{app:more-details}
\label{app:protocol}

\subsection{Quantizer and Cache Accounting}
\label{app:quantizer-accounting}

All cached tensors use asymmetric uniform quantization with FP16 metadata. A
group of values $x$ with minimum $m$ and maximum $M$ is quantized to $b$ bits
as
\begin{equation}
s=\frac{M-m}{2^{b}-1},\qquad
q=\mathrm{clamp}\!\left(\mathrm{round}\!\left(\frac{x-m}{s}\right),\,0,\,
2^{b}-1\right),\qquad
\hat{x}=s\,q+m,
\end{equation}
where $s$ and $m$ are stored in FP16 and rounding is to nearest, with ties to
even. The two cached tensors differ in bit width, grouping, and clipping
(Table~\ref{tab:app-quantizer}). The offline fit uses the same quantizers as
the runtime, matched bit for bit.

\begin{table}[h]
\centering
\caption{Quantizer configuration of RAQ-MLA. $P$ is the cache-page size in
tokens.}
\label{tab:app-quantizer}
\small
\begin{tabular}{@{}lll@{}}
\toprule
 & Latent $\mathbf{c}^{\mathrm{KV}}$ & Positional key $\mathbf{k}^{\mathrm{R}}$ \\
\midrule
Quantized tensor & $\mathbf{c}^{\mathrm{KV}}_tA_{\mathrm{c}}$ & $\mathbf{k}^{\mathrm{R}}_tA_{\mathrm{R}}$ (after RoPE) \\
Bits (levels) & $2$ ($4$) & $4$ ($16$) \\
Grouping & per token, groups of $128$ channels & per channel, within a cache page \\
Clipping & per token, $0.94$ quantile of $|x|$ & none \\
Metadata & FP16 scale and minimum per group & FP16 scale and minimum per channel and page \\
Effective bits & $2+32/128=2.25$ & $4+32/P$ ($4.25$ at $P=128$, $4.50$ at $P=64$) \\
\bottomrule
\end{tabular}
\end{table}

\paragraph{Latent.}
The latent is quantized per token in its learned basis. Before grouping, each
token's $d_{\mathrm{c}}=512$ channels are clipped: with $\tau_t$ the $94$th percentile of
the token's channel magnitudes, every channel is clamped to
$[-\tau_t,\tau_t]$, which affects the $30$ largest of the $512$ channels. At
two bits a group has only four levels, so a few large channels would otherwise
stretch its range and coarsen the step for all others; clipping trades their
error for a finer step on the rest. The clipped row is then split into four
groups of $128$ contiguous channels, each quantized with its own scale and
minimum.

\paragraph{Positional key.}
The positional key is quantized after RoPE and $A_{\mathrm{R}}$, per channel within each
cache page: the $P$ values of one channel in one page share a scale and a
minimum. This grouping follows the fixed-channel outliers of $\mathbf{k}^{\mathrm{R}}$
(Section~\ref{sec:motivation}): a large channel receives its own range rather
than stretching the range of the other channels in the same token. No clipping
is applied.

\paragraph{Cache pages and the BF16 tail.}
Following KVarN \citep{kvarn}, a page is quantized when it is sealed. New
tokens are written to the open page in BF16, and the whole page is quantized
once its last slot is written; per-channel quantization of $\mathbf{k}^{\mathrm{R}}$
requires this anyway, because its scales are computed over the tokens of a
page. The page size follows the attention backend: $P=128$ tokens for
GLM-4.7-Flash, Moonlight, and DS-V2-Lite, and $P=64$ for GLM-5.2, whose
sparse-attention backend uses $64$-token pages. The most recent tokens
therefore stay in BF16, but on a sawtooth schedule rather than as a sliding
window: the BF16 region grows to a full page and then collapses. All quantized
methods, including the baselines, use the same page schedule.

\paragraph{Bits per element.}
We report effective bits per cached element (BPE), counting code bits and FP16
metadata and averaging over both cached tensors,
\begin{equation}
\mathrm{BPE}=\frac{d_{\mathrm{c}}\,b_{\mathrm{c}}+d^{\mathrm{R}}\,b_{\mathrm{R}}}{d_{\mathrm{c}}+d^{\mathrm{R}}},
\end{equation}
where $b_{\mathrm{c}}$ and $b_{\mathrm{R}}$ are the effective bits of the latent and the positional
key. RAQ-MLA costs $(512\cdot2.25+64\cdot4.25)/576=2.47$ BPE with $128$-token
pages and $2.50$ with $64$-token pages, a $6.47\times$ and $6.40\times$
capacity increase over BF16. Methods that leave the positional key in BF16
cost at least $(512\cdot2.25+64\cdot16)/576=3.78$ BPE ($4.23\times$) even with
a two-bit latent. The RULER results in Table~\ref{tab:ruler} raise the first
two and last two layers to an INT4 latent and an INT8 positional key, which
costs $2.66$ BPE over the $47$ layers of GLM-4.7-Flash. BPE excludes the BF16
open page, which holds at most one page per sequence and layer; at $64$K tokens
it adds at most $0.2\%$ of the tokens, or $0.03$ bits per element.

\subsection{Evaluation Protocol}
\label{app:evaluation-protocol}

All models are served with vLLM behind an OpenAI-compatible API, with the
quantizers of Appendix~\ref{app:quantizer-accounting} applied when a cache page
is written. Table~\ref{tab:app-serving} lists the serving and sampling
configuration of the reasoning and coding tasks; within a model, all methods
share it.

\begin{table}[h]
\centering
\caption{Serving and sampling configuration for the reasoning and coding tasks.
ARC-Challenge, PIQA, and RULER NIAH use greedy decoding instead.}
\label{tab:app-serving}
\small
\begin{tabular}{@{}lcccc@{}}
\toprule
 & GLM-4.7-Flash & GLM-5.2-FP8 & Moonlight & DS-V2-Lite \\
\midrule
\texttt{max\_model\_len} & $202{,}752$ & $202{,}752$ & $8{,}192$ & $32{,}768$ \\
\texttt{max\_tokens} & $131{,}072$ & $65{,}536$ & $7{,}168$ & $7{,}168$ \\
Cache page (tokens) & $128$ & $64$ & $128$ & $128$ \\
Temperature & $1.0$ & $1.0$ & $1.0$ & $1.0$ \\
Top-$p$ & $0.95$ & $0.95$ & $0.95$ & $0.95$ \\
\bottomrule
\end{tabular}
\end{table}

AIME25, GPQA-Diamond, MATH-500, GSM8K, and HumanEval are evaluated with
EvalScope 1.9.1, using eight concurrent requests, the sampling settings of
Table~\ref{tab:app-serving}, and a generation budget of \texttt{max\_tokens}.
We report accuracy, and pass@1 for HumanEval. Because sampling makes individual
runs noisy, some of the reported results are averaged over three seeds.
ARC-Challenge and PIQA, used for Moonlight and DS-V2-Lite, are scored as
multiple-choice questions with greedy decoding at temperature $0$. RULER NIAH
is evaluated with OpenCompass over all eight needle subtasks at $4$K, $8$K,
$16$K, $32$K, $64$K, and $128$K, with $100$ samples per subtask and length,
greedy decoding, an output limit of $128$ tokens, and thinking disabled.

\section{Extended Results}
\label{app:full-results}

\paragraph{Long-context understanding.}
Table~\ref{tab:app-longbench-groups} reports the LongBench-v1 macro average
over all $21$ tasks on GLM-4.7-Flash, with thinking enabled. RAQ-MLA stores a
$30\%$ smaller cache than OSCAR and scores on par with it.

\begin{table}[!ht]
\centering
\caption{\textbf{LongBench-v1 macro average (\%) over $21$ tasks on
GLM-4.7-Flash}, thinking on.}
\label{tab:app-longbench-groups}
\small
\setlength{\tabcolsep}{3pt}
\resizebox{\columnwidth}{!}{%
\begin{tabular}{lcccccccc}
\toprule
Method & BPE & Single-QA & Multi-QA & Summ. & Few-shot & Synth. & Code & Avg \\
\midrule
OSCAR & 3.78 & \textbf{49.16} & \textbf{57.24} & \textbf{22.88} & 60.46
 & \textbf{77.20} & \textbf{54.75} & \textbf{52.38} \\
\ours \textbf{RAQ-MLA} & \textbf{2.66} & 49.07 & 56.83 & 22.80
 & \textbf{61.84} & 76.30 & 52.44 & 52.19 \\
\bottomrule
\end{tabular}%
}
\end{table}

\paragraph{Decode latency across output lengths.}
Table~\ref{tab:app-output-sweep} extends the single operating point of
Table~\ref{tab:packed-runtime-latency} to output lengths from $1$K to $8$K
tokens, with a $512$-token prompt and batch size $1$. The packed cache lowers
TPOT by $6.8$--$9.0\%$ at every length as the cache grows to $8.7$K tokens.
The gain is smaller than the $1.38\times$ of the decode kernel in
Table~\ref{tab:packed-runtime-latency} because attention is only part of each
decode step at this scale.

\begin{table}[!ht]
\centering
\caption{\textbf{Decode TPOT across output lengths on GLM-4.7-Flash}, with a
$512$-token prompt and batch size $1$. Speedup is the FlashMLA TPOT divided by
the RAQ-MLA TPOT.}
\label{tab:app-output-sweep}
\small
\begin{tabular}{rrrr}
\toprule
Output tokens & FlashMLA (ms) & RAQ-MLA (ms) & Speedup \\
\midrule
$1024$ & $6.61$ & $\mathbf{6.04}$ & $1.09\times$ \\
$2048$ & $6.69$ & $\mathbf{6.11}$ & $1.09\times$ \\
$4096$ & $6.87$ & $\mathbf{6.25}$ & $1.10\times$ \\
$8192$ & $7.02$ & $\mathbf{6.54}$ & $1.07\times$ \\
\bottomrule
\end{tabular}
\end{table}

\paragraph{Cost of the latent transform.}
Applying $A_{\mathrm{c}}$ is the only per-token matrix product on the latent
write path. We measure it on one H20 GPU with a single MLA layer that has the
cache shape of GLM-4.7-Flash ($d_{\mathrm{c}}=512$, $d^{\mathrm{R}}=64$),
$64$-token pages, and $2$K cached tokens, under CUDA Graphs
(Table~\ref{tab:app-transform-cost}). The layer has $64$ heads; the Hopper
kernel processes heads in tiles of $64$, so this costs the same as the $10$
heads per GPU of GLM-4.7-Flash. Fused into the cache-write kernel, the
transform costs $1.7$--$2.4$~$\mu$s per layer for decode batches from $1$ to
$64$, nearly independent of the batch size. Over $47$ layers this is about
$0.1$~ms per decode step, under $2\%$ of the $6$~ms TPOT at batch size $1$.
Its share of the per-layer write-plus-attention time falls from $9.9\%$ at
batch size $1$ to $1.0\%$ at batch size $64$, since attention grows with the
batch while the transform does not. Each layer stores one $512$~KiB BF16
matrix, $23.5$~MiB per GPU over $47$ layers, and the fused and unfused paths
write byte-identical cache entries.

\begin{table}[!ht]
\centering
\caption{\textbf{Per-layer cost of the latent transform} on a single
GLM-shaped MLA layer with $2$K cached tokens. Standalone applies $A_{\mathrm{c}}$ as a
separate GEMM before the cache write; fused applies it inside the write kernel.
Share is the fused cost relative to the per-layer cache write plus attention.}
\label{tab:app-transform-cost}
\small
\begin{tabular}{rrrr}
\toprule
Decode batch & Standalone ($\mu$s) & Fused ($\mu$s) & Share \\
\midrule
$1$  & $2.66$ & $\mathbf{2.36}$ & $9.9\%$ \\
$2$  & $2.68$ & $\mathbf{1.72}$ & $6.6\%$ \\
$4$  & $2.64$ & $\mathbf{1.77}$ & $5.5\%$ \\
$8$  & $2.70$ & $\mathbf{1.81}$ & $4.0\%$ \\
$64$ & $2.83$ & $\mathbf{2.24}$ & $1.0\%$ \\
\bottomrule
\end{tabular}
\end{table}

\end{document}